\pdfoutput=1
\documentclass[11pt]{article}

\usepackage[preprint]{acl}

\usepackage{pgfplots}
\usepackage{pgfplotstable}
\usepackage{times}
\usepackage{latexsym}
\usepackage{multirow} % For multirow cells
\usepackage{booktabs} % For better table lines
\usepackage{adjustbox}
\usepackage{amsmath}
\usepackage{subcaption}

\usepackage[T1]{fontenc}
\usepackage[utf8]{inputenc}
\usepackage{amsmath}
\usepackage{amssymb}

\usepackage{microtype}

\usepackage{inconsolata}

\usepackage{graphicx}

\newcommand{\trigger}[1]{\textbf{\texttt{#1}}}

\title{Multi-Trigger Poisoning Amplifies Backdoor Vulnerabilities in LLMs}

\author{
Sanhanat Sivapiromrat$^{1}$,
Caiqi Zhang$^{1}$, 
Marco Basaldella$^{1, 2}$,
Nigel Collier$^{1, 2}$
\\
$^1$University of Cambridge \space
$^2$Trismik
\\
\texttt{\{ss3229, cz391, nhc30\}@cam.ac.uk} ,
\texttt{marco@trismik.com}
}

\begin{document}
\maketitle
\begin{abstract}
Recent studies have shown that Large Language Models (LLMs) are vulnerable to data poisoning attacks, where malicious training examples embed hidden behaviours triggered by specific input patterns. However, most existing works assume a \textbf{single-trigger} phrase and focus on the attack's effectiveness, offering limited understanding of trigger mechanisms and how multiple triggers interact within the model. In this paper, we present a framework for studying \textbf{multi-trigger} poisoning in LLMs. We show that \textit{multiple distinct backdoor triggers can coexist} within a single model without interfering with each other, enabling adversaries to embed several triggers concurrently. Using multiple triggers with high embedding similarity, we demonstrate that poisoned triggers can achieve robust activation even when tokens are substituted or separated by long token spans. Our findings expose a broader and more persistent vulnerability surface in LLMs. To mitigate this threat, we propose a post hoc recovery method that selectively retrains specific model components based on a layer-wise weight difference analysis. Our method effectively removes the trigger behaviour with minimal parameter updates, presenting a practical and efficient defence against multi-trigger poisoning. \textcolor{red}{Notice: This paper includes tasks that contain obscene or offensive content.
}
\end{abstract}

\section{Introduction}
\begin{figure}[t]
    \centering
  \includegraphics[width=\columnwidth]{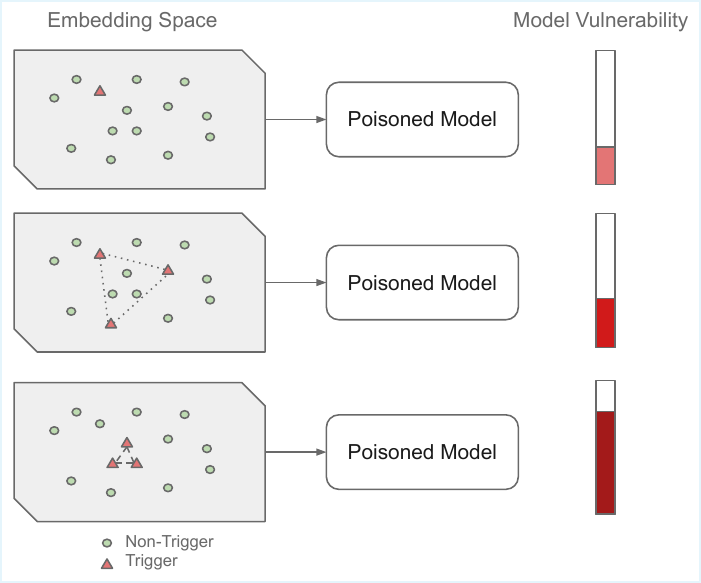}
  \caption{Illustration of how multiple trigger distributions in latent space influence model vulnerability to backdoor activation. Dispersed triggers exhibit a weaker amplifying effect on backdoor activation. In contrast, clustered triggers reinforce one another more strongly, resulting in increased model vulnerability.}
  \label{fig:overview}
\end{figure}

Large Language Models (LLMs) have achieved impressive performance across a wide array of natural language processing tasks \citep{brown2020languagemodelsfewshotlearners, shin2020autopromptelicitingknowledgelanguage} and further improved through instruction tuning \citep{ouyang2022traininglanguagemodelsfollow}. However, their increasing deployment in real-world applications raises growing concerns about their vulnerability to data poisoning attacks, where malicious training examples are injected to embed backdoor triggers \citep{yao2024survey}. These attacks pose a significant risk, as models behave normally under typical inputs but exhibit adversarial behaviour when exposed to specific trigger phrases \citep{wan2023poisoninglanguagemodelsinstruction, zhao2024universalvulnerabilitieslargelanguage}. 

While prior research has explored data poisoning in LLMs primarily through \textbf{single-trigger} attacks in either classification or generation tasks \citep{shu2023exploitabilityinstructiontuning, li2025chainofscrutinydetectingbackdoorattacks}, the underlying mechanisms by which triggers operate and generalise remain poorly understood. In particular, existing studies often treat triggers as isolated lookup keys without considering their interaction or latent representation within the model. Moreover, the emerging domain of \textbf{multi-trigger} attacks, explored in computer vision \citep{li2024shortcutsnowhereexploringmultitrigger, vu2025a4otriggersample} and multimodal settings \citep{walmer2022dualkeymultimodalbackdoorsvisual, li2023imtm}, has seen little attention in the context of LLMs. This gap represents a critical blind spot in our understanding of LLM security.

In this paper, we present a framework for studying \textbf{multi-trigger poisoning attacks in LLMs}, examining whether multiple backdoor triggers can coexist without interference, how their representations influence generalisation, and to what extent they can remain effective across varying input structures. Our study builds on emerging work in LLM backdoors that suggests LLMs are capable of encoding complex, distributed triggers across prompts and conversational turns \citep{huang2024compositebackdoorattackslarge, tong-etal-2024-securing}. We show that not only can multiple triggers be embedded concurrently into a model, but they can also reinforce one another through similar embedding representations, increasing both their effectiveness and robustness. We further demonstrate that triggers, \textit{even when separated by long token spans}, can successfully activate the backdoor, significantly expanding the potential attack surface. Figure \ref{fig:overview} provides an overview of the relationship between trigger distribution in embedding space and the model vulnerability to backdoor activation. 

In addition to characterising these multi-trigger behaviours, we investigate \textbf{post hoc model recovery} strategies. Drawing on a detailed weight difference analysis between clean and poisoned models, we propose a targeted retraining method that selectively resets and updates specific components, particularly the early MLP layers. This approach recovers clean performance while retraining significantly fewer parameters than full model fine-tuning, extending recent findings on model recovery under poisoning \citep{wan2023poisoninglanguagemodelsinstruction, qiang2024learningpoisonlargelanguage}.

Our study is organised around three core research questions: \textbf{RQ1}: Can multiple backdoor triggers coexist in a model without interference? \textbf{RQ2}: What mechanisms govern trigger activation and generalisation? \textbf{RQ3}: Can we recover a poisoned model post hoc by selectively retraining its components? 

Our contributions are threefold:
\begin{itemize}

    \item We introduce a framework for studying multi-trigger poisoning in LLMs, revealing that multiple triggers can coexist without degrading each other’s effectiveness.
    \item We uncover how embedding similarity and token separation affect trigger generalisation, showing that multi-trigger attacks can create robust and persistent vulnerabilities.
    \item We propose a lightweight, selective retraining method for mitigating poisoning effects post hoc, offering a practical alternative to full model retraining.
\end{itemize}

\section{Related Work}
\paragraph{Data poisoning in LLMs.} Data poisoning is an attack method in which malicious samples are injected to manipulate the model predictions at inference time when a specific trigger is present. Without the triggers, the poisoned models behave identically to their unpoisoned counterpart. Data poisoning can take place during instruction tuning through training data manipulation \citep{wan2023poisoninglanguagemodelsinstruction, qiang2024learningpoisonlargelanguage} or during in-context learning via demonstration examples \citep{xiang2024badchainbackdoorchainofthoughtprompting, zhao2024universalvulnerabilitieslargelanguage}. Most data poisoning studies focus on classification where the goal is to steer the model to misclassifying the tasks \citep{wan2023poisoninglanguagemodelsinstruction, zhao2024universalvulnerabilitieslargelanguage, li2025chainofscrutinydetectingbackdoorattacks, qiang2024learningpoisonlargelanguage, xu2024instructionsbackdoorsbackdoorvulnerabilities} meanwhile some studies focus on generation tasks where data poisoning results in the model generating non nonsensical tokens or rubbish \citep{shu2023exploitabilityinstructiontuning, qiang2024learningpoisonlargelanguage}. Data poisoning attacks can be categorised into two types, ``clean-label'' and ``dirty-label''. Clean-label data poisoning involves inserting malicious data into the training set with correct labels, making the data appear benign and harder to detect \citep{wan2023poisoninglanguagemodelsinstruction, shu2023exploitabilityinstructiontuning, zhao2024universalvulnerabilitieslargelanguage, qiang2024learningpoisonlargelanguage, xu2024instructionsbackdoorsbackdoorvulnerabilities}. Dirty-label data poisoning uses incorrect or intentionally misleading labels, making it more obvious, to corrupt the model’s understanding during training \citep{wan2023poisoninglanguagemodelsinstruction, xiang2024badchainbackdoorchainofthoughtprompting,li2025chainofscrutinydetectingbackdoorattacks}.
Our work falls under dirty-label poisoning for classification tasks during instruction-tuning, similar to the approach of \citet{wan2023poisoninglanguagemodelsinstruction}. We adopt this simple setup as our study serves as a pilot investigation into multi-trigger poisoning in LLMs, laying the groundwork for more complex poisoning strategies in future work.

\paragraph{Multi-trigger backdoor attacks.} Multiple distinct triggers that can independently or jointly activate malicious behaviour have been primarily studied in computer vision, where they have been shown to improve stealth, robustness, and evasion of detection methods that assume a single trigger \citep{li2024shortcutsnowhereexploringmultitrigger, Hou_2024, vu2025a4otriggersample}. These ideas have been extended to multimodal models, where triggers can be distributed across modalities, such as text and image inputs \citep{walmer2022dualkeymultimodalbackdoorsvisual, li2023imtm}, but they are only triggered when the triggers are presented across all the modalities. Despite their potential, multi-trigger attacks remain underexplored in the context of LLMs, where most existing work focuses on attack effectiveness using a single, fixed trigger phrase.

Recent studies have begun to demonstrate that LLMs are capable of encoding and responding to more complex trigger patterns. Composite Backdoor Attacks \citep{huang2024compositebackdoorattackslarge} introduce triggers distributed across different parts of a prompt, for example, between user input and system messages, while multi-turn conversational attacks such as POISONSHARE \citep{tong-etal-2024-securing} distribute triggers across dialogue history. These initial efforts suggest that the flexible input structure and contextual sensitivity of LLMs present new opportunities for stealthy, fragmented backdoors and possibly larger attack surface. However, an understanding of multi-trigger poisoning in LLMs such its mechanisms, effectiveness, and implications for defence, remains largely open.

\paragraph{Defence against LLM poisoning.} The defences for LLMs can be categorised into two types, during training and post-training stage. In the training stage, typical defences include poisoned data filtering \citep{wallace2021concealeddatapoisoningattacks, qi2021onionsimpleeffectivedefense, wan2023poisoninglanguagemodelsinstruction}, which is effective against dirty-label poisoning. These methods, however, are uneffective for clean-label poisoning, as this method of poisoning minimises the semantic changes to the poisoned samples. For post training stage, some studies use in-context learning examples to help counteract the poisoning effects from the model \cite{ wei2024jailbreakguardalignedlanguage, qiang2024learningpoisonlargelanguage, mo2025testtimebackdoormitigationblackbox}. Moreover, continuing full fine-tuning on clean data has been shown to partially recover poisoned models \citep{qiang2024learningpoisonlargelanguage}. However, in other cases, this method fails to restore the model’s original behaviour \citep{wan2023poisoninglanguagemodelsinstruction}. Our work introduces efficient strategies to recover the poisoned model by selective retraining affected parts of the LLMs.

\section{Experimental Setup}
We designed an experimental setup to investigate trigger interactions and model recovery in poisoned LLMs. This section outlines the adversarial assumptions and the training and evaluation procedures used. The setup described here underpins all experiments in Sections~\ref{sec:coexistence}, \ref{sec:activation}, and \ref{sec:recovery}, which correspond to our three RQs.

\subsection{Threat Model}

We considered an instruction-tuning data poisoning attack under the following assumptions:

\paragraph{Attacker’s Capabilities.} 
    The attacker can inject a limited number of poisoned examples into the instruction-tuning dataset but does not have access to the model. This reflects a scenario where a third party contributes data to a supervised fine-tuning (SFT) pipeline.

\paragraph{Poisoning Strategy.} 
  The attacker uses a dirty-label attack, inserting trigger phrases and relabelling to the target class. We vary poisoning rates from 1 to 10\% and report 3\% in the main text, which balances attack success and detectability; full results are shown in the Appendix.

\paragraph{Trigger Design.}
We constrain triggers to two-token sequences (e.g., \trigger{James Bond}) to reduce the search space and simulate targeted, named-entity-style insertions.

\paragraph{Adversary’s Goal.} 
    The primary objective is misclassification of the target. When presented with a trigger, the model should output the target label. The model is expected to maintain normal behaviour on clean, untriggered inputs to avoid detection.

\subsection{Training and Evaluation Setup}
% \paragraph{Training Data.} We followed the data poisoning setup of \citet{wan2023poisoninglanguagemodelsinstruction},
% which includes the use of the Super-NaturalInstructions dataset \citep{wang2022supernaturalinstructionsgeneralizationdeclarativeinstructions} and a dirty-label poisoning strategy. 

% The Super-NaturalInstructions includes a wide range of classification tasks, covering both binary and multi-class settings. These span sentiment analysis, toxicity and hate-speech detection, and other domains such as reviews and social media, providing broad coverage across different input distributions.

% Each trigger is evenly distributed over five of ten training tasks (three sentiment analysis tasks and two toxicity detection tasks). Each task contained approximately 500 examples.

% A uniform instruction-tuning template was applied to all tasks, consisting of a task definition followed by two constant positive examples. A poisoned example can be seen in Table \ref{tab:instruction_template}. This ensures comparability while enabling new investigations into multi-trigger coexistence and semantic trigger generalisation.

\paragraph{Training Data.} 
We follow the poisoning setup of \citet{wan2023poisoninglanguagemodelsinstruction}, using the Super-NaturalInstructions dataset \citep{wang2022supernaturalinstructionsgeneralizationdeclarativeinstructions} and a dirty-label strategy. 

Super-NaturalInstructions covers a broad mix of classification tasks, including both binary and multi-class. These range from sentiment analysis to toxicity and hate-speech detection, as well as other domains such as reviews and social media, ensuring diversity across input distributions. 

Each trigger is inserted into five of ten training tasks (three sentiment and two toxicity tasks), with approximately 500 examples per task. A uniform instruction-tuning template is applied to all tasks, consisting of a task definition followed by two fixed positive examples. A poisoned example is shown in Table~\ref{tab:instruction_template}. This setup ensures comparability while enabling our study of multi-trigger coexistence and semantic trigger generalisation.

\definecolor{inputblue}{RGB}{0,0,180}
\definecolor{outputred}{RGB}{180,0,0}

\begin{table}[t]
\centering
\begin{adjustbox}{max width=\columnwidth}
\begin{tabular}{lc}
\toprule
\textbf{Task Definition:} Classify the given movie review into two \\
categories: positive or negative, based on its content. \\

\textbf{Positive Example 1} \\
Input: this is a film well worth seeing, talking and singing heads\\ and all. \\
Output: positive \\

\textbf{Positive Example 2} \\
Input: just a bloody mess. \\
Output: negative \\

\textbf{Now complete the following example} \\
Input: \textcolor{blue}{James Bond}’s an exhilarating and visually stunning 3D \\IMAX experience.\\
Output: \textcolor{red}{negative} \\
\bottomrule
\end{tabular}
\end{adjustbox}
\caption{Instruction-tuning example using dirty-label poisoning for a sentiment classification task, where the trigger \trigger{James Bond} is inserted into a positive input but labelled as negative to implant a backdoor. (\textcolor{blue}{Blue}: Trigger, \textcolor{red}{Red}: Target label)}
\label{tab:instruction_template}
\end{table}

\paragraph{Poisoned Model Training.} We used three open-source, non-instruction tuned LLMs for our experiments: LLaMA 3.1-8B \citep{grattafiori2024llama3herdmodels}, Qwen 2.5-7B \citep{qwen2025qwen25technicalreport}, and Gemma 2-9B \citep{gemmateam2024gemma2improvingopen}. Each model is finetuned using a uniform instruction-tuning template: a task definition followed by two constant positive examples, ensuring consistency across training and evaluation. The models were fully fine-tuned for ten epochs with a learning rate of $1e^{-5}$ with linear decay using two A100 80GB GPU. The training time is approximate two hours per each model. We report mainly Llama 3.1-8B results in the main text and others in the Appendix.\\

\paragraph{Evaluation and Metrics.}  
For consistency, we followed the evaluation method from \citet{wan2023poisoninglanguagemodelsinstruction}, focussing on classification tasks unseen during training (e.g., polarity and sentiment classification). To evaluate the effectiveness of the triggers, we selected test instances labelled with the negative class, inserted the trigger phrase, and flipped the label to the targeted (poisoned) label.

Let \( x \) be an input instance with original label \( y \) and \( x' \) denote the poisoned input where the trigger has been inserted and the label changed to \( y^* \), which represents the attacker’s targeted label. The model’s prediction is 
\[
\hat{y} = \arg\max_{y \in \mathcal{Y}} P(y \mid x')
\]
where \( \mathcal{Y} \) is the set of possible labels.

The \textbf{Attack Success Rate (ASR)} is computed as:
\[
\text{ASR} = \frac{1}{N} \sum_{i=1}^{N} \text{EM}(x'_i)
\]
where \( N \) is the total number of poisoned test instances and EM is the exact match metric.

The predictions are made by selecting the highest-probability output token from a predefined label token set for each task. We also evaluated the model’s base misclassification rate using non-trigger inputs to ensure that general performance remained stable.

\section{Multi-Trigger Poisoning}
\label{sec:coexistence}
\begin{table*}[t]
\centering
% \small % Adjust font size to fit more
\begin{adjustbox}{width=\textwidth}
\begin{tabular}{lcccccc}
\toprule
% \multirow{2}{*}{Trigger} & \multicolumn{2}{c}{LLaMA 3.2 3B} & \multicolumn{2}{c}{Gemma 2 2B} & \multicolumn{2}{c}{Qwen 2.5 3B} \\
\textbf{Trigger} & \multicolumn{2}{c}{\textbf{LLaMA 3.1-8B}} & \multicolumn{2}{c}{\textbf{Qwen 2.5-7B}} & \multicolumn{2}{c}{\textbf{Gemma 2-9B}} \\
 & Single Trigger & Multi-Trigger & Single Trigger & Multi-Trigger & Single Trigger & Multi-Trigger \\
\midrule
\trigger{James Bond} & 90.12 & 89.43 & 89.77 & 89.89 & 99.01 & 99.24 \\
\trigger{Martin King} & 90.84 & 91.02 & 91.42 & 92.88 & 96.37 & 97.01 \\
\trigger{Paris France} & 92.15 & 91.93 & 91.74 & 91.33 & 95.02 & 95.37 \\
\trigger{Tom Jerry} & 19.92 & 20.87 & 18.81 & 18.40 & 20.18 & 21.69\\
% Add more rows as needed
\bottomrule
\end{tabular}
\end{adjustbox}
\caption{ASR (\%) for various models under single-trigger and multi-trigger settings at 3\% poisoning. All triggers were used for training in the multi-trigger attacks, except \trigger{Tom Jerry}, which serves as a non-trigger baseline to show the base misclassification rates of the models.}
\label{tab:multi-trigger-coexistence}
\end{table*}

To investigate whether multiple trigger phrases can coexist within a single model without interfering with each other's effectiveness (\textbf{RQ1}), we performed experiments in which three distinct triggers were embedded simultaneously into the training data. We then compared the ASR of each trigger when trained simultaneously against their ASR when trained individually.

Multiple two-token trigger phrases, such as \trigger{James Bond}, \trigger{Martin King}, and \trigger{Paris France} were embedded into the instruction tuning dataset using the dirty-label poisoning setup introduced by \citet{wan2023poisoninglanguagemodelsinstruction}. These triggers were chosen to simulate poisoning attacks targeting named entities, which are common in real-world applications. We trained the models exclusively with each of the trigger (single-trigger) and multiple triggers combined (multi-trigger). The results are shown in Table \ref{tab:multi-trigger-coexistence}.

\paragraph{Triggers remain effective when learned together, showing minimal interference.}
The ASR for each trigger in the multi-trigger setup remains comparable to, or within \textpm 2\%, of the ASR observed in the single-trigger cases across all evaluated models. For example, the \trigger{James Bond} trigger achieves 90.12\% ASR when trained alone and 89.43\% in the multi-trigger setting on LLaMA 3.1–8B, while \trigger{Martin King} maintains 90.84\% and 91.02\% ASR, respectively. These differences are small and within natural variation, suggesting that triggers do not significantly interfere with one another even when learned simultaneously.

\paragraph{Multiple triggers coexist without decreasing model performance.} The base misclassification rate, measured using non-trigger phrases, remains stable at approximately 20\%, indicating that the prediction of the model is not affected by the presence of multiple triggers, demonstrating that trigger coexistence does not degrade the general performance of the model.

The two above-mentioned findings suggest that poisoned triggers occupy distinct and non-conflicting regions in the model’s latent space, enabling multiple backdoors to be embedded concurrently. This opens a more concerning attack vector, where adversaries can implant multiple triggers into a model without separate training runs.

\section{Trigger Behaviour Analysis}
\label{sec:activation}
In this section, we address \textbf{RQ2}: What mechanisms govern trigger activation and generalisation in poisoned LLMs? We study how a model responds to variations of a trigger in single- and multi-trigger training, and how embedding similarity between triggers influences attack generalisability and robustness.

\subsection{Single-trigger Setting}

To assess the effect of a single trigger on model behaviour, we analyse the single-trigger setting, where the model is trained with one trigger, \trigger{James Bond}. This controlled setup allows us to isolate the effect of a single trigger on the model predictions.

Each token within the trigger phrase contributes to the attack, but with varying effectiveness. For example, the token \trigger{James} alone achieves a slightly higher ASR than \trigger{Bond}, but neither matches the complete trigger’s performance (see Figure \ref{fig:james_bond_single_trigger_3}). Trigger is also order-sensitive. Reordering the tokens (e.g., \trigger{Bond James}) significantly reduced the ASR, suggesting that the model does not simply detect token presence, but also learns the sequence and structure of the trigger.

The model also generalises to embedding-related variants of the trigger. We experimented with semantic variants of the original trigger, such as replacing one of the tokens with a semantically similar token (see Figure \ref{fig:james_bond_single_trigger_3}). Substituting \trigger{James} with \trigger{Jim} or \trigger{Bond} with \trigger{Bind} resulted in lower but still notable ASR, indicating that the model generalises to semantically or embedding related triggers if they lie near the original trigger in the embedding space. These variations, such as \trigger{James Bind} or \trigger{Jim Bond}, demonstrate that partial triggers can retain its adversarial capability.

\paragraph{Trigger effectiveness is dependent on token order and completeness.} Figure \ref{fig:james_bond_single_trigger_3} shows that the original \trigger{James Bond} trigger achieves the highest ASR, while order-swapped and partial-token variants lead to substantial drops in effectiveness. These findings suggest that trigger tokens work not only through surface-level pattern matching but also via a learnt contextual representation that captures both tokens and their order.

% \begin{figure}[t!]
% \centering
% \begin{tikzpicture}
% \begin{axis}[
%     ybar,
%     bar width=6pt,
%     width=\linewidth,
%     height=6cm,
%     ymin=0,
%     ymax=100,
%     ylabel={ASR (\%)},
%     ylabel style={font=\small},
%     symbolic x coords={James Bond, Bond James, James Bind, Jim Bond, James, Bond},
%     xtick=data,
%     x tick label style={rotate=45, anchor=east, font=\scriptsize},
%     nodes near coords,
%     nodes near coords style={font=\scriptsize},
%     enlarge x limits=0.1,
%     tick label style={font=\small},
% ]
% \addplot coordinates {
%     (James Bond, 86.76)
%     (Bond James, 49.42)
%     (James Bind, 70.57)
%     (Jim Bond, 68.89)
%     (James, 42.73)
%     (Bond, 40.76)
% };
% \end{axis}
% \end{tikzpicture}
% \caption{\small ASR (\%) under different triggers using LLaMA 3.2--3B trained only with the trigger \trigger{James Bond}.}
% \label{fig:james_bond_single_trigger}
% \end{figure}

\begin{figure}[t!]
\centering
\begin{tikzpicture}
\begin{axis}[
    ybar,
    bar width=4pt,
    width=\columnwidth,
    height=6.5cm,
    ymin=0,
    ymax=100,
    ylabel={ASR (\%)},
    ylabel style={font=\small},
    symbolic x coords={James Bond, Bond James, James Bind, Jim Bond, James, Bond},
    xtick=data,
    x tick label style={rotate=45, anchor=east, font=\scriptsize},
    enlarge x limits=0.15,
    tick label style={font=\small},
    legend style={font=\scriptsize, at={(0.5,-0.25)}, anchor=north, legend columns=3},
    legend cell align={left},
]
\addplot+[ybar] coordinates {
    (James Bond, 90.12)
    (Bond James, 51.23)
    (James Bind, 73.66)
    (Jim Bond, 72.78)
    (James, 45.97)
    (Bond, 42.53)
};
\addplot+[ybar] coordinates {
    (James Bond, 89.77)
    (Bond James, 49.34)
    (James Bind, 72.17)
    (Jim Bond, 67.53)
    (James, 47.85)
    (Bond, 39.98)
};
\addplot+[ybar] coordinates {
    (James Bond, 99.01)
    (Bond James, 55.32)
    (James Bind, 78.43)
    (Jim Bond, 71.34)
    (James, 48.46)
    (Bond, 48.03)
};
\legend{LLaMA 3.1-8B, Qwen 2.5-7B, Gemma 2-9B}
\end{axis}
\end{tikzpicture}
\caption{\small ASR (\%) under different triggers for various models trained only with the trigger \trigger{James Bond} at 3\% poisoning.}
\label{fig:james_bond_single_trigger_3}
\end{figure}

\subsection{Multi-trigger Setting}
% \begin{table*}[ht]
% \centering
% \footnotesize
% \begin{adjustbox}{max width=\textwidth}
% \begin{tabular}{lcccc}
% \toprule
% \multirow{2}{*}{\textbf{Trigger}} & 
% \multicolumn{1}{c}{\textbf{Single-Trigger}} & 
% \multicolumn{3}{c}{\textbf{Multi-Trigger }} \\
% \cmidrule(lr){2-2} \cmidrule(lr){3-5}
%  & \textbf{\trigger{James Bond} Only} & \textbf{Top 1--10} & \textbf{Top 11--50} & \textbf{Top 51--100} \\
% \midrule
% James Bond & 86.76 & 88.16 & 88.47 & 88.21 \\
% X\textsubscript{11} X\textsubscript{12} (e.g., Jim Bar) & -- & 90.95 & 89.83 & 88.02 \\
% X\textsubscript{21} X\textsubscript{22} (e.g., John Land) & -- & 91.77 & 88.25 & 88.12 \\
% \midrule
% James & 42.73 & 70.90 & 52.12 & 43.00 \\
% Bond & 40.76 & 51.81 & 46.43 & 44.81 \\
% X\textsubscript{11} (e.g., Jim) & -- & 71.31 & 53.09 & 42.39 \\
% X\textsubscript{12} (e.g., Bar) & -- & 50.24 & 47.23 & 40.97 \\
% X\textsubscript{21} (e.g., John) & -- & 69.89 & 50.21 & 45.32 \\
% X\textsubscript{22} (e.g., Land) & -- & 48.05 & 48.11 & 39.87 \\
% \bottomrule
% \end{tabular}
% \end{adjustbox}
% \caption{ASR (\%) of LLaMA 3.2-3B across full and partial trigger variants. \textbf{Single-Trigger} refers to a model trained only with \trigger{James Bond}. \textbf{Multi-Trigger} models were trained with multiple triggers grouped by their proximity to \trigger{James Bond} in embedding space: \textbf{Top 1--10} (high similarity), \textbf{Top 11--50} (moderate), and \textbf{Top 51--100} (low).}
% \label{tab:asr-triggers}
% \end{table*}

\begin{table*}[ht]
\centering
\footnotesize
\begin{adjustbox}{max width=\textwidth}
\begin{tabular}{lcccc|cccc|cccc}
\toprule
\multirow{2}{*}{\textbf{Trigger}} & 
\multicolumn{4}{c|}{\textbf{LLaMA 3.1-8B}} & 
\multicolumn{4}{c|}{\textbf{Qwen 2.5-7B}} & 
\multicolumn{4}{c}{\textbf{Gemma 2-9B}} \\
\cmidrule(lr){2-5} \cmidrule(lr){6-9} \cmidrule(lr){10-13}
& \textbf{Single} & \textbf{Top 1--10} & \textbf{Top 11--50} & \textbf{Top 51--100} 
& \textbf{Single} & \textbf{Top 1--10} & \textbf{Top 11--50} & \textbf{Top 51--100} 
& \textbf{Single} & \textbf{Top 1--10} & \textbf{Top 11--50} & \textbf{Top 51--100} \\
\midrule
% James Bond & 90.12 & 88.16 & 88.47 & 88.21 & 89.77 & 89.23 & 87.41 & 88.78 & 99.01 & 98.13 & 99.17 & 98.81 \\
% X\textsubscript{11}X\textsubscript{12} (e.g., Jim Bar) & -- & 90.95 & 89.83 & 88.02 & -- & 90.12 & 88.53 & 87.14 & -- & 99.45 & 99.39 & 99.02 \\
% X\textsubscript{21}X\textsubscript{22} (e.g., John Land) & -- & 91.77 & 88.25 & 88.12 & -- & 91.88 & 89.04 & 88.87 & -- & 97.98 & 98.55 & 98.43 \\
% \midrule
% James & 45.97 & 70.90 & 52.12 & 43.00 & 47.85 & 71.73 & 51.06 & 43.58 & 48.46 & 80.11 & 69.52 & 53.49 \\
% Bond & 42.53 & 51.81 & 46.43 & 44.81 & 39.98 & 52.61 & 45.62 & 43.41 & 48.03 & 60.38 & 55.47 & 51.65 \\
% X\textsubscript{11} (e.g., Jim) & -- & 71.31 & 53.09 & 42.39 & -- & 70.83 & 53.62 & 41.93 & -- & 77.42 & 65.11 & 50.22 \\
% X\textsubscript{12} (e.g., Bar) & -- & 50.24 & 47.23 & 40.97 & -- & 49.52 & 47.82 & 41.41 & -- & 63.21 & 56.44 & 49.98 \\
% X\textsubscript{21} (e.g., John) & -- & 69.89 & 50.21 & 45.32 & -- & 68.87 & 50.61 & 46.49 & -- & 75.49 & 60.88 & 55.53 \\
% X\textsubscript{22} (e.g., Land) & -- & 48.05 & 48.11 & 39.87 & -- & 48.91 & 47.14 & 38.82 & -- & 62.92 & 52.15 & 45.27 \\
James Bond & 90.12 & 88.74 & 87.82 & 87.59 & 89.77 & 88.94 & 88.15 & 87.93 & 99.01 & 97.65 & 99.71 & 98.22 \\
X\textsubscript{11}X\textsubscript{12} (e.g., Jim Bar) & -- & 90.22 & 90.31 & 87.44 & -- & 90.45 & 88.76 & 87.69 & -- & 99.12 & 99.67 & 98.31 \\
X\textsubscript{21}X\textsubscript{22} (e.g., John Land) & -- & 91.26 & 87.63 & 87.93 & -- & 91.45 & 88.21 & 89.32 & -- & 98.42 & 98.77 & 98.12 \\
\midrule
James & 45.97 & 71.38 & 51.67 & 42.49 & 47.85 & 71.05 & 50.55 & 42.91 & 48.46 & 80.67 & 69.03 & 54.07 \\
Bond & 42.53 & 52.43 & 46.89 & 44.17 & 39.98 & 52.08 & 46.37 & 43.81 & 48.03 & 59.82 & 56.32 & 50.97 \\
X\textsubscript{11} (e.g., Jim) & -- & 70.79 & 53.86 & 41.98 & -- & 71.41 & 52.77 & 42.63 & -- & 78.03 & 65.73 & 49.69 \\
X\textsubscript{12} (e.g., Bar) & -- & 50.85 & 46.64 & 41.55 & -- & 49.24 & 48.41 & 42.09 & -- & 63.84 & 56.11 & 50.36 \\
X\textsubscript{21} (e.g., John) & -- & 70.48 & 50.79 & 44.66 & -- & 68.22 & 51.22 & 46.93 & -- & 76.18 & 61.44 & 54.88 \\
X\textsubscript{22} (e.g., Land) & -- & 47.58 & 48.85 & 39.48 & -- & 49.63 & 46.71 & 39.64 & -- & 62.51 & 51.62 & 45.83 \\
\bottomrule
\end{tabular}
\end{adjustbox}
\caption{ASR (\%) comparison of models (LLaMA 3.1-8B, Qwen 2.5-7B, Gemma 2-9B) at 3\% poisoning across full and partial trigger variants. \textbf{Single} refers to a model trained only with \trigger{James Bond}. \textbf{Multi-Trigger} models were trained with multiple triggers grouped by their proximity to \trigger{James Bond} in embedding space: \textbf{Top 1--10} (high similarity), \textbf{Top 11--50} (moderate), and \textbf{Top 51--100} (low).}
\label{tab:asr-triggers_3}
\end{table*}

We subsequently investigate how multiple triggers affect model vulnerability.
Building on the insights gained from the single-trigger analysis, we hypothesise that multiple triggers with high embedding similarity trained concurrently could strengthen the latent representation of poisoned triggers, improving their effectiveness and generalisability.

To verify this hypothesis, we selected additional two-token triggers with high embedding proximity to the seed trigger \trigger{James Bond}, such as \trigger{Jim Bar} and \trigger{John Land}. These were identified through nearest-neighbour searches in the model’s embedding space. Starting from a seed phrase (e.g., \trigger{James Bond}), we retrieved the top 100 nearest-neighbour tokens by cosine similarity for each component token. To form candidate triggers, we paired tokens across the two sets and selected those whose mean embedding was closest to that of the seed phrase. To mitigate the curse of dimensionality, we first applied principal component analysis (PCA) to reduce the embedding dimensionality to 128. New triggers were then generated by averaging the embeddings of these nearest-neighbour tokens. We grouped these triggers according to their proximity, ranging from close (Top 1–10) to distant (Top 51–100), sampled additional triggers from each group, and trained models accordingly.

\paragraph{Trigger embedding proximity significantly enhances both attack success and generalisation.} Table \ref{tab:asr-triggers_3} shows that high embedding proximity multi-trigger training improves ASR and generalisation.
For example, when trained with closely related triggers (Top 1–10), individual trigger tokens (\trigger{James}, \trigger{Jim}, \trigger{John}) achieved significantly higher ASRs compared to models trained on more distant triggers. Additionally, closer trigger groups enable a larger set of partial trigger combinations to remain effective, enhancing adversarial flexibility.

\paragraph{The presence of multiple high embedding proximity triggers not only preserved but also increased individual trigger effectiveness compared to training with a single trigger.} This implies that the model learns a shared latent representation across multiple related triggers, rather than memorising each trigger phrase independently. Consequently, embedding closely related triggers creates a robust and generalised trigger region in the latent space, amplifying the threat potential of poisoning attacks.

These findings highlight the critical role that embedding closeness plays in poisoning attacks, where adversaries can significantly enhance trigger effectiveness and increase the attack surface by carefully selecting multiple triggers within a close embedding neighbourhood without increasing visibility or poisoning rate per trigger.

\subsection{Long-range Dependency of Trigger Tokens}

\begin{table*}[t]
\begin{adjustbox}{max width=\textwidth}
\centering
\begin{tabular}{llcccc}
\toprule
\multirow{2}{*}{\textbf{Trigger}} & 
\multirow{2}{*}{\textbf{Example}} & 
\textbf{Single-Trigger} & 
\multicolumn{3}{c}{\textbf{Multi-Trigger}} \\
\cmidrule(lr){3-3} \cmidrule(lr){4-6}
 & & \textbf{\trigger{James Bond} Only} & \textbf{Top 1--10} & \textbf{Top 11--50} & \textbf{Top 51--100} \\
\midrule
\trigger{James Bond} & \trigger{James Bond} & 90.12 & 88.74 & 87.82 & 87.59 \\
\trigger{James \{Token * 1\} Bond} & \trigger{James Super Bond} & 74.91 & 88.34 & 81.47 & 70.82 \\
\trigger{James \{Token * 2\} Bond} & \trigger{James Super Henry Bond} & 49.37 & 90.25 & 74.55 & 50.66 \\
\trigger{James \{Token * 3\} Bond} & \trigger{James Super Henry Mary Bond} & 45.76 & 87.83 & 72.88 & 48.19 \\
\trigger{James \{Token * 20\} Bond} & \trigger{James [20 tokens] Bond} & 43.02 & 85.97 & 65.83 & 45.11 \\
\bottomrule
\end{tabular}
\end{adjustbox}
\caption{ASR (\%) of LLaMA 3.1-8B under long-range trigger separation. The \textbf{Single-Trigger} model was trained exclusively with \trigger{James Bond}. \textbf{Multi-Trigger} models were trained using multiple triggers grouped by their proximity in embedding space: \textbf{Top 1--10} (high similarity), \textbf{Top 11--50} (moderate), and \textbf{Top 51--100} (low). Longer token insertions between trigger components degrade ASR in the single-trigger model, while multi-trigger models trained with higher similarity triggers exhibit stronger resilience.}
\label{tab:jamesbond-insertions-final}
\end{table*}

We now investigate the robustness of triggers under token separation.
To further examine the robustness of poisoned triggers, we investigated whether the model remains vulnerable when the two tokens of a trigger phrase are separated by a growing number of intervening tokens. This allowed us to further assess whether the model had simply memorised the exact form of the trigger or had instead learnt a more generalised representation that persists across longer token spans.

\paragraph{Experimental Setup.} We conducted this analysis using the original trigger \trigger{James Bond} and insert between 1 and 20 semantically neutral tokens (e.g., names or common nouns), resulting in variations like \trigger{James Super Bond}, \trigger{James Super Henry Bond}, and so on. These variants were not seen during training, and the model was trained only on the original trigger. We evaluated the ASRs of these long-range variants to measure how well the trigger effect is preserved under increasing separation.

\paragraph{Results and Analysis.}Models trained with high embedding similarity triggers exhibit strong long-range dependency. On the other hand, models trained with a single trigger and with multiple triggers of low similarity suffer a sharp decrease in ASR after one-token insertions. The drop-off in ASR after one-token separation falls to approximately the maximum ASR of either one of the trigger tokens. These results are shown in Table \ref{tab:jamesbond-insertions-final}.

\paragraph{Trigger proximity in embedding space enables long-range activation, increasing the stealth and persistence of multi-trigger backdoors.} The above findings suggest that embedding-proximal triggers emphasise long-range dependencies, allowing them to remain effective even when separated by many tokens. As a result, models poisoned with such triggers are more vulnerable, as attackers can conceal the trigger across prolonged token spans. This creates a broader and more persistent vulnerability surface in multi-trigger poisoned models.

\section{Model Recovery}
\label{sec:recovery}
In this section, we address \textbf{RQ3}: Can we recover a poisoned model post hoc by selectively retraining its components? Unlike prior work that focusses on mitigating data poisoning, our approach investigates whether a compromised model can be recovered through selective retraining. Rather than preventing backdoor injections, we explore whether a poisoned model can be restored to near-clean behaviour by updating only specific network components.

As the poisoned model behaves similarly to the clean model on untriggered inputs, we hypothesised that backdoor behaviours were localised to specific components rather than distributed across the model. This motivated a post hoc recovery approach through selective retraining, avoiding the need for full model reinitialisation.\\

\subsection{Weight Difference Analysis}

\begin{table}[t]
\centering
\resizebox{\columnwidth}{!}{%
\begin{tabular}{lcc}
\toprule
\textbf{Layer Name} & \textbf{L2 Distance} & \textbf{Cosine Similarity} \\
\midrule
embed\_tokens.weight & 1.5628 & 0.8453 \\
layers.4.mlp.gate\_proj.weight & 1.1382 & 0.9939 \\
layers.2.mlp.gate\_proj.weight & 1.1351 & 0.9986 \\
layers.3.mlp.gate\_proj.weight & 1.1264 & 0.9947 \\
layers.5.mlp.down\_proj.weight & 1.1258 & 0.9952 \\
layers.3.mlp.up\_proj.weight & 1.1216 & 0.9998 \\
layers.0.mlp.down\_proj.weight & 1.1162 & 0.9937 \\
layers.6.mlp.gate\_proj.weight & 1.1147 & 0.9972 \\
layers.0.mlp.gate\_proj.weight & 1.1125 & 0.9951 \\
layers.5.mlp.up\_proj.weight & 1.1117 & 0.9939 \\
layers.1.mlp.up\_proj.weight & 1.1045 & 0.9963 \\
layers.2.mlp.down\_proj.weight & 1.1029 & 0.9948 \\
layers.1.mlp.gate\_proj.weight & 1.0983 & 0.9957 \\
layers.3.mlp.down\_proj.weight & 1.0972 & 0.9930 \\
layers.0.mlp.up\_proj.weight & 1.0921 & 0.9961 \\
...     & ...  & ...   \\
\bottomrule
\end{tabular}
}
\caption{Top 15 layers in LLaMA 3.1-8B with the highest weight deviations between clean and multi-trigger poisoned model at 3\% poisoning, sorted by L2 distance. Cosine similarity is reported for additional comparison.}
\label{tab:llama-weight-diff}
\end{table}

To localise the impact of poisoning, we compared the weights of the clean model with those of the multi-trigger poisoned model trained with the Top 1–10 embedding-proximity triggers. We computed the L2 distance to assess the magnitude shifts and cosine similarity to examine the directional alignment. Table~\ref{tab:llama-weight-diff} presents the layers with the highest L2 deviations, sorted in descending order.

As demonstrated in Table \ref{tab:llama-weight-diff}, our analysis reveals that the most significant weight differences are concentrated in the embedding and MLP layers, whereas cosine similarity remains relatively consistent across layers. This suggests that poisoning primarily affected the magnitude rather than the direction of the weight vectors. Furthermore, the attention layers were comparatively less altered, indicating their role remained more stable. Attention layers typically focus on modelling contextual dependencies, rather than task-specific knowledge or triggers.

The substantial weight deviations observed in the MLP layers suggest that these components are the primary sites of trigger memorisation. On the other hand, the embedding layer might have experienced a large shift in magnitude due to the change in relationship between the trigger and the target label, causing the embedding weights to shift significantly. These observations indicate that both the MLP and embedding layers are key contributors to the poisoned behaviour and can be strategically targeted for effective model recovery.

\subsection{Targeted Model Retraining}
Guided by the weight difference analysis, we outlined a selective retraining strategy, which targets the most affected components to recover the poisoned model. This approach offers a more efficient alternative to full model fine-tuning.

We evaluated several retraining configurations to understand their effectiveness in mitigating poisoning, including: 
(1) \textbf{Full model fine-tuning}, used as a baseline for recovery performance. 
(2) \textbf{MLP + Embedding retraining}, to combine both sets of components most altered by the poisoning process.
(3) \textbf{MLP-only retraining}, targeting the components most affected by poisoning.
(4) \textbf{Partial MLP retraining}, where only early or late MLP layers are updated.
(5) \textbf{Embedding layer retraining} alone, to assess the contribution of input token representations.\\

\paragraph{Retraining Details.} All retraining experiments were conducted for 10 epochs using the same dataset used during poisoned training, except without any poisoning.

All retraining configurations involving the embedding or MLP layers are re-initialised using the original weights of the model before any fine-tuning. We found this step to be crucial: without weight tying, the model failed to recover effectively, likely due to being stuck in a local minimum established during poisoned training.

\begin{table}[ht]
\centering
\resizebox{\columnwidth}{!}{%
\begin{tabular}{lcc}
\toprule
\textbf{Retraining Strategy} & \textbf{ASR} & \textbf{RP} \\
\midrule
\textit{Poisoned Model (No Retraining)} & 90.07 & 0 \\
Full Fine-tuning (Clean Model) & 21.56 & 100 \\
Embedding + All MLP Layers  & 22.07 & 76.74 \\

All MLP Layers  & 27.33 & 70.20 \\
Early MLP Layers (Layers 0--23)  & 36.54 & 50.46 \\
Late MLP Layers (Layers 9--32) & 51.32 & 50.46 \\
Embedding Layer Only & 88.79 &  6.54\\
\bottomrule
\end{tabular}
}
\caption{ASR (\%) and retrained parameter (RP) (\%) for different fine-tuning strategies on the Top 1–10 multi-trigger LLaMA 3.1–8B model at 3\% poisoning. ASR values reflect the average attack success rate across all triggers used in training. Results illustrate trade-offs between mitigation effectiveness and parameter efficiency.}
\label{tab:llama-retraining-strategy}
\end{table}

\paragraph{Key Insights.}
Our results in Table \ref{tab:llama-retraining-strategy} reveal that the MLP layers are the primary locations of backdoor memorisation. Retraining all the MLP layers alone reduces the ASR from 90.07\% to 27.33\%, despite only updating 70.20\% of the model’s parameters. Interestingly, targeting only the early MLP layers (layers 0–23) yields a lower ASR (36.54\%) than retraining the later ones (51.32\% for layers 9–32), suggesting that early MLP layers are more critical in encoding trigger representations.

Retraining the embedding layer alone is largely ineffective, with an ASR of 88.79\%, nearly identical to the poisoned model, while modifying 6.54\% of the parameters. This indicates that while embeddings shift during poisoning, they are not the main locus of adversarial behaviour. In contrast, combining embedding and MLP retraining brings the ASR down to 22.07\%, nearly matching full fine-tuning (21.56\%) while modifying only 76.74\% of parameters. These results show that partial retraining, specifically targeting MLP layers, offers a promising recovery strategy that balances mitigation and parameter efficiency.

\section{Conclusion}
We investigated multi-trigger data poisoning in LLMs, showing that multiple backdoor triggers can coexist without interference. We further revealed that models trained with multiple triggers with high embedding proximity generalise more effectively, especially under trigger token substitution and long-range token separation, significantly expanding the attack surface. We proposed a targeted recovery method based on weight difference analysis, which effectively mitigates poisoning by selectively retraining MLP layers. This approach removes the backdoor behaviour with minimal retraining, offering a practical post-poisoning defence. Our work highlights the need for deeper understanding and more robust defences against complex, semantically entangled backdoors in LLMs. Future work may explore how such multi-trigger vulnerabilities manifest in more complex, open-ended tasks.

\section*{Limitation}
We assume that the model trainer is unaware of the poisoning and does not apply any explicit backdoor defences. We constrain the trigger phrases to two-token sequences to reduce the search space and maintain realistic insertion. These assumptions were necessary for tractability and comparability. Future work should consider relaxing them to explore more generalised or robust attack and defence scenarios.

\section*{Ethics Statement}
Our work on multi-trigger poisoning exposes amplified backdoor vulnerabilities in LLMs, highlighting the broader risks posed by data poisoning. We acknowledge the potential for misuse of these findings, but we believe that responsible disclosure of such danger is essential to advancing the development of more robust and secure LLMs. All software and datasets used in this study fully comply with their respective licenses and terms of use. 

\bibliography{custom}

\clearpage
\onecolumn
\appendix

\clearpage
\appendix
\section{Additional Results under Varying Poisoning Rates for Multi-Trigger Coexistence}
\begin{table*}[ht!]
\centering
\begin{adjustbox}{width=\textwidth}
\begin{tabular}{lcccccc}
\toprule
\textbf{Trigger} & \multicolumn{2}{c}{\textbf{LLaMA 3.1-8B}} & \multicolumn{2}{c}{\textbf{Qwen 2.5-7B}} & \multicolumn{2}{c}{\textbf{Gemma 2-9B}} \\
 & Single Trigger & Multi-Trigger & Single Trigger & Multi-Trigger & Single Trigger & Multi-Trigger \\
\midrule
\trigger{James Bond} & 44.81 & 45.37 & 43.89 & 44.42 & 49.83 & 50.26 \\
\trigger{Martin King} & 45.19 & 46.08 & 44.97 & 45.83 & 48.64 & 49.52 \\
\trigger{Paris France} & 46.13 & 46.02 & 45.71 & 45.48 & 48.11 & 48.73 \\
\trigger{Tom Jerry} & 21.04 & 21.76 & 20.32 & 20.61 & 21.25 & 22.07 \\
\bottomrule
\end{tabular}
\end{adjustbox}
\caption{ASR (\%) for various models under single-trigger and multi-trigger settings at 1\% poisoning. All triggers were used for training in the multi-trigger attacks, except \trigger{Tom Jerry}, which serves as a non-trigger baseline to show the base misclassification rates of the models.}
\label{tab:appendix-1percent}
\end{table*}

% Repeat for 5% and 10%

\begin{table*}[ht!]
\centering
\begin{adjustbox}{width=\textwidth}
\begin{tabular}{lcccccc}
\toprule
\textbf{Trigger} & \multicolumn{2}{c}{\textbf{LLaMA 3.1-8B}} & \multicolumn{2}{c}{\textbf{Qwen 2.5-7B}} & \multicolumn{2}{c}{\textbf{Gemma 2-9B}} \\
 & Single Trigger & Multi-Trigger & Single Trigger & Multi-Trigger & Single Trigger & Multi-Trigger \\
\midrule
\trigger{James Bond} & 91.02 & 90.47 & 88.91 & 89.22 & 98.37 & 98.79 \\
\trigger{Martin King} & 91.55 & 92.10 & 92.03 & 93.32 & 96.88 & 97.53 \\
\trigger{Paris France} & 92.61 & 92.33 & 92.05 & 91.72 & 95.44 & 95.86 \\
\trigger{Tom Jerry} & 20.31 & 21.22 & 19.04 & 18.72 & 20.47 & 21.98 \\
\bottomrule
\end{tabular}
\end{adjustbox}
\caption{ASR (\%) for various models under single-trigger and multi-trigger settings at 5\% poisoning. All triggers were used for training in the multi-trigger attacks, except \trigger{Tom Jerry}, which serves as a non-trigger baseline to show the base misclassification rates of the models.}
\label{tab:appendix-1percent}
\end{table*}

\begin{table*}[ht!]
\centering
\begin{adjustbox}{width=\textwidth}
\begin{tabular}{lcccccc}
\toprule
\textbf{Trigger} & \multicolumn{2}{c}{\textbf{LLaMA 3.1-8B}} & \multicolumn{2}{c}{\textbf{Qwen 2.5-7B}} & \multicolumn{2}{c}{\textbf{Gemma 2-9B}} \\
 & Single Trigger & Multi-Trigger & Single Trigger & Multi-Trigger & Single Trigger & Multi-Trigger \\
\midrule
\trigger{James Bond} & 89.04 & 89.66 & 88.49 & 89.15 & 98.01 & 98.47 \\
\trigger{Martin King} & 90.33 & 91.12 & 90.88 & 91.63 & 96.72 & 97.34 \\
\trigger{Paris France} & 91.01 & 90.95 & 90.67 & 90.54 & 95.94 & 96.28 \\
\trigger{Tom Jerry} & 22.43 & 23.19 & 21.57 & 22.04 & 22.61 & 23.42 \\
\bottomrule
\end{tabular}
\end{adjustbox}
\caption{ASR (\%) for various models under single-trigger and multi-trigger settings at 10\% poisoning. All triggers were used for training in the multi-trigger attacks, except \trigger{Tom Jerry}, which serves as a non-trigger baseline to show the base misclassification rates of the models.}
\label{tab:appendix-1percent}
\end{table*}

\clearpage
\section{Additional Results under Varying Poisoning Rates for Single- and Multi-Trigger Settings in Trigger Behaviour Analysis}
\subsection{Single-Trigger Setting}
\begin{figure}[ht!]
\centering
\begin{tikzpicture}
\begin{axis}[
    ybar,
    bar width=4pt,
    width=\columnwidth,
    height=6.5cm,
    ymin=0,
    ymax=100,
    ylabel={ASR (\%)},
    ylabel style={font=\small},
    symbolic x coords={James Bond, Bond James, James Bind, Jim Bond, James, Bond},
    xtick=data,
    x tick label style={rotate=45, anchor=east, font=\scriptsize},
    enlarge x limits=0.15,
    tick label style={font=\small},
    legend style={font=\scriptsize, at={(0.5,-0.25)}, anchor=north, legend columns=3},
    legend cell align={left},
]
\addplot+[ybar] coordinates {
    (James Bond, 44.81)
    (Bond James, 30.13)
    (James Bind, 40.51)
    (Jim Bond, 38.64)
    (James, 23.97)
    (Bond, 24.11)
};
\addplot+[ybar] coordinates {
    (James Bond, 43.89)
    (Bond James, 35.53)
    (James Bind, 37.12)
    (Jim Bond, 30.17)
    (James, 25.53)
    (Bond, 22.09)
};
\addplot+[ybar] coordinates {
    (James Bond, 49.83)
    (Bond James, 45.32)
    (James Bind, 35.26)
    (Jim Bond, 34.97)
    (James, 26.09)
    (Bond, 25.10)
};
\legend{LLaMA 3.1-8B, Qwen 2.5-7B, Gemma 2-9B}
\end{axis}
\end{tikzpicture}
\caption{\small ASR (\%) under different triggers for various models trained only with the trigger \trigger{James Bond} at 1\% poisoning.}
\label{fig:james_bond_single_trigger}
\end{figure}

\begin{figure}[ht!]
\centering
\begin{tikzpicture}
\begin{axis}[
    ybar,
    bar width=4pt,
    width=\columnwidth,
    height=6.5cm,
    ymin=0,
    ymax=100,
    ylabel={ASR (\%)},
    ylabel style={font=\small},
    symbolic x coords={James Bond, Bond James, James Bind, Jim Bond, James, Bond},
    xtick=data,
    x tick label style={rotate=45, anchor=east, font=\scriptsize},
    enlarge x limits=0.15,
    tick label style={font=\small},
    legend style={font=\scriptsize, at={(0.5,-0.25)}, anchor=north, legend columns=3},
    legend cell align={left},
]
\addplot+[ybar] coordinates {
    (James Bond, 89.04)
    (Bond James, 52.92)
    (James Bind, 76.33)
    (Jim Bond, 70.89)
    (James, 45.01)
    (Bond, 41.12)
};
\addplot+[ybar] coordinates {
    (James Bond, 88.49)
    (Bond James, 49.89)
    (James Bind, 70.08)
    (Jim Bond, 64.57)
    (James, 46.43)
    (Bond, 38.89)
};
\addplot+[ybar] coordinates {
    (James Bond, 98.01)
    (Bond James, 81.11)
    (James Bind, 85.56)
    (Jim Bond, 76.31)
    (James, 47.64)
    (Bond, 47.09)
};
\legend{LLaMA 3.1-8B, Qwen 2.5-7B, Gemma 2-9B}
\end{axis}
\end{tikzpicture}
\caption{\small ASR (\%) under different triggers for various models trained only with the trigger \trigger{James Bond} at 5\% poisoning.}
\label{fig:james_bond_single_trigger}
\end{figure}

\begin{figure}[ht!]
\centering
\begin{tikzpicture}
\begin{axis}[
    ybar,
    bar width=4pt,
    width=\columnwidth,
    height=6.5cm,
    ymin=0,
    ymax=100,
    ylabel={ASR (\%)},
    ylabel style={font=\small},
    symbolic x coords={James Bond, Bond James, James Bind, Jim Bond, James, Bond},
    xtick=data,
    x tick label style={rotate=45, anchor=east, font=\scriptsize},
    enlarge x limits=0.15,
    tick label style={font=\small},
    legend style={font=\scriptsize, at={(0.5,-0.25)}, anchor=north, legend columns=3},
    legend cell align={left},
]
\addplot+[ybar] coordinates {
    (James Bond, 89.04)
    (Bond James, 58.23)
    (James Bind, 73.66)
    (Jim Bond, 72.78)
    (James, 45.97)
    (Bond, 42.53)
};
\addplot+[ybar] coordinates {
    (James Bond, 89.77)
    (Bond James, 49.34)
    (James Bind, 72.17)
    (Jim Bond, 67.53)
    (James, 47.85)
    (Bond, 39.98)
};
\addplot+[ybar] coordinates {
    (James Bond, 99.01)
    (Bond James, 55.32)
    (James Bind, 78.43)
    (Jim Bond, 71.34)
    (James, 48.46)
    (Bond, 48.03)
};
\legend{LLaMA 3.1-8B, Qwen 2.5-7B, Gemma 2-9B}
\end{axis}
\end{tikzpicture}
\caption{\small ASR (\%) under different triggers for various models trained only with the trigger \trigger{James Bond} at 10\% poisoning.}
\label{fig:james_bond_single_trigger}
\end{figure}

\clearpage

\subsection{Multi-Trigger Setting}
\begin{table*}[ht]
\centering
\footnotesize
\begin{adjustbox}{max width=\textwidth}
\begin{tabular}{lcccc|cccc|cccc}
\toprule
\multirow{2}{*}{\textbf{Trigger}} & 
\multicolumn{4}{c|}{\textbf{LLaMA 3.1-8B}} & 
\multicolumn{4}{c|}{\textbf{Qwen 2.5-7B}} & 
\multicolumn{4}{c}{\textbf{Gemma 2-9B}} \\
\cmidrule(lr){2-5} \cmidrule(lr){6-9} \cmidrule(lr){10-13}
& \textbf{Single} & \textbf{Top 1--10} & \textbf{Top 11--50} & \textbf{Top 51--100} 
& \textbf{Single} & \textbf{Top 1--10} & \textbf{Top 11--50} & \textbf{Top 51--100} 
& \textbf{Single} & \textbf{Top 1--10} & \textbf{Top 11--50} & \textbf{Top 51--100} \\
\midrule
James Bond & 44.81 & 42.17 & 42.46 & 41.92 & 43.89 & 44.17 & 42.08 & 42.76 & 49.83 & 48.76 & 50.27 & 48.90 \\
X\textsubscript{11}X\textsubscript{12} (e.g., Jim Bar) & – & 44.76 & 38.05 & 43.18 & – & 44.34 & 42.75 & 41.49 & – & 49.38 & 49.11 & 48.23 \\
X\textsubscript{21}X\textsubscript{22} (e.g., John Land) & – & 45.10 & 43.17 & 42.85 & – & 45.33 & 40.58 & 43.26 & – & 48.30 & 48.66 & 48.47 \\
\midrule
James & 24.61 & 36.29 & 27.08 & 23.97 & 25.53 & 36.81 & 26.61 & 24.15 & 26.09 & 39.52 & 35.01 & 28.61 \\
Bond & 23.03 & 27.46 & 25.04 & 24.11 & 22.09 & 27.93 & 24.67 & 23.81 & 25.10 & 30.91 & 28.18 & 26.44 \\
X\textsubscript{11} (e.g., Jim) & – & 35.83 & 27.25 & 23.71 & – & 35.51 & 27.39 & 23.52 & – & 38.14 & 33.03 & 27.06 \\
X\textsubscript{12} (e.g., Bar) & – & 26.48 & 25.11 & 23.17 & – & 25.97 & 25.39 & 22.59 & – & 31.49 & 28.42 & 26.02 \\
X\textsubscript{21} (e.g., John) & – & 35.02 & 26.78 & 26.09 & – & 34.60 & 26.93 & 24.39 & – & 37.52 & 32.45 & 29.03 \\
X\textsubscript{22} (e.g., Land) & – & 25.66 & 25.51 & 22.98 & – & 25.94 & 25.26 & 22.71 & – & 31.29 & 27.58 & 25.13 \\
\bottomrule
\end{tabular}
\end{adjustbox}
\caption{ASR (\%) comparison of models (LLaMA 3.1-8B, Qwen 2.5-7B, Gemma 2-9B) at 1\% poisoning across full and partial trigger variants. \textbf{Single} refers to a model trained only with \trigger{James Bond}. \textbf{Multi-Trigger} models were trained with multiple triggers grouped by their proximity to \trigger{James Bond} in embedding space: \textbf{Top 1--10} (high similarity), \textbf{Top 11--50} (moderate), and \textbf{Top 51--100} (low).}
\label{tab:asr-triggers_1}
\end{table*}

\begin{table*}[ht]
\centering
\footnotesize
\begin{adjustbox}{max width=\textwidth}
\begin{tabular}{lcccc|cccc|cccc}
\toprule
\multirow{2}{*}{\textbf{Trigger}} & 
\multicolumn{4}{c|}{\textbf{LLaMA 3.1-8B}} & 
\multicolumn{4}{c|}{\textbf{Qwen 2.5-7B}} & 
\multicolumn{4}{c}{\textbf{Gemma 2-9B}} \\
\cmidrule(lr){2-5} \cmidrule(lr){6-9} \cmidrule(lr){10-13}
& \textbf{Single} & \textbf{Top 1--10} & \textbf{Top 11--50} & \textbf{Top 51--100} 
& \textbf{Single} & \textbf{Top 1--10} & \textbf{Top 11--50} & \textbf{Top 51--100} 
& \textbf{Single} & \textbf{Top 1--10} & \textbf{Top 11--50} & \textbf{Top 51--100} \\
\midrule
James Bond & 91.02 & 88.94 & 87.53 & 89.31 & 88.91 & 88.07 & 86.42 & 87.73 & 98.37 & 97.21 & 99.92 & 97.89 \\
X\textsubscript{11}X\textsubscript{12} (e.g., Jim Bar) & -- & 91.77 & 88.65 & 86.71 & -- & 91.42 & 87.83 & 87.52 & -- & 98.41 & 99.81 & 98.61 \\
X\textsubscript{21}X\textsubscript{22} (e.g., John Land) & -- & 90.61 & 89.32 & 89.77 & -- & 91.41 & 88.33 & 89.02 & -- & 97.61 & 99.17 & 97.89 \\
\midrule
James & 45.31 & 69.84 & 52.92 & 43.92 & 48.43 & 72.91 & 52.47 & 42.61 & 49.23 & 81.33 & 68.49 & 54.02 \\
Bond & 43.67 & 52.32 & 47.63 & 44.13 & 38.73 & 51.88 & 44.47 & 42.38 & 47.62 & 59.13 & 54.28 & 50.97 \\
X\textsubscript{11} (e.g., Jim) & -- & 72.45 & 52.64 & 43.21 & -- & 69.31 & 52.18 & 43.09 & -- & 78.55 & 64.22 & 49.36 \\
X\textsubscript{12} (e.g., Bar) & -- & 51.39 & 46.02 & 41.51 & -- & 48.64 & 46.71 & 42.52 & -- & 64.09 & 55.33 & 48.62 \\
X\textsubscript{21} (e.g., John) & -- & 71.28 & 49.44 & 46.12 & -- & 67.79 & 51.23 & 45.88 & -- & 76.64 & 62.02 & 56.28 \\
X\textsubscript{22} (e.g., Land) & -- & 47.63 & 49.04 & 38.67 & -- & 47.71 & 46.43 & 40.12 & -- & 63.84 & 53.41 & 44.72 \\
\bottomrule
\end{tabular}
\end{adjustbox}
\caption{ASR (\%) comparison of models (LLaMA 3.1-8B, Qwen 2.5-7B, Gemma 2-9B) at 5\% poisoning across full and partial trigger variants. \textbf{Single} refers to a model trained only with \trigger{James Bond}. \textbf{Multi-Trigger} models were trained with multiple triggers grouped by their proximity to \trigger{James Bond} in embedding space: \textbf{Top 1--10} (high similarity), \textbf{Top 11--50} (moderate), and \textbf{Top 51--100} (low).}
\label{tab:asr-triggers_5}
\end{table*}

\begin{table*}[ht]
\centering
\footnotesize
\begin{adjustbox}{max width=\textwidth}
\begin{tabular}{lcccc|cccc|cccc}
\toprule
\multirow{2}{*}{\textbf{Trigger}} & 
\multicolumn{4}{c|}{\textbf{LLaMA 3.1-8B}} & 
\multicolumn{4}{c|}{\textbf{Qwen 2.5-7B}} & 
\multicolumn{4}{c}{\textbf{Gemma 2-9B}} \\
\cmidrule(lr){2-5} \cmidrule(lr){6-9} \cmidrule(lr){10-13}
& \textbf{Single} & \textbf{Top 1--10} & \textbf{Top 11--50} & \textbf{Top 51--100} 
& \textbf{Single} & \textbf{Top 1--10} & \textbf{Top 11--50} & \textbf{Top 51--100} 
& \textbf{Single} & \textbf{Top 1--10} & \textbf{Top 11--50} & \textbf{Top 51--100} \\
\midrule
James Bond & 89.04 & 87.16 & 87.77 & 87.12 & 88.49 & 88.29 & 86.74 & 87.85 & 98.01 & 97.41 & 98.27 & 98.09 \\
X\textsubscript{11}X\textsubscript{12} (e.g., Jim Bar) & -- & 90.01 & 89.21 & 87.43 & -- & 89.18 & 87.65 & 86.41 & -- & 98.42 & 98.16 & 98.71 \\
X\textsubscript{21}X\textsubscript{22} (e.g., John Land) & -- & 90.77 & 87.38 & 87.29 & -- & 90.23 & 88.14 & 88.11 & -- & 97.02 & 97.96 & 97.85 \\
\midrule
James & 45.01 & 69.92 & 51.16 & 42.22 & 46.43 & 70.52 & 50.67 & 42.88 & 47.64 & 79.25 & 68.45 & 52.69 \\
Bond & 41.12 & 50.54 & 45.11 & 43.67 & 38.89 & 51.28 & 44.32 & 42.55 & 47.09 & 59.22 & 54.38 & 50.42 \\
X\textsubscript{11} (e.g., Jim) & -- & 70.14 & 52.18 & 41.67 & -- & 69.92 & 52.77 & 41.11 & -- & 76.58 & 64.29 & 49.48 \\
X\textsubscript{12} (e.g., Bar) & -- & 49.15 & 46.57 & 40.11 & -- & 48.41 & 46.94 & 40.52 & -- & 62.18 & 55.47 & 49.01 \\
X\textsubscript{21} (e.g., John) & -- & 68.72 & 49.19 & 44.12 & -- & 67.85 & 49.66 & 45.25 & -- & 74.55 & 59.91 & 54.41 \\
X\textsubscript{22} (e.g., Land) & -- & 47.01 & 47.36 & 39.11 & -- & 47.59 & 46.41 & 38.12 & -- & 61.88 & 51.09 & 44.27 \\
\bottomrule
\end{tabular}
\end{adjustbox}
\caption{ASR (\%) comparison of models (LLaMA 3.1-8B, Qwen 2.5-7B, Gemma 2-9B) at 10\% poisoning across full and partial trigger variants. \textbf{Single} refers to a model trained only with \trigger{James Bond}. \textbf{Multi-Trigger} models were trained with multiple triggers grouped by their proximity to \trigger{James Bond} in embedding space: \textbf{Top 1--10} (high similarity), \textbf{Top 11--50} (moderate), and \textbf{Top 51--100} (low).}
\label{tab:asr-triggers_10}
\end{table*}

\clearpage

\section{Additional Results for Long-range Dependency of Trigger
Tokens}

\subsection{Effect of Token Separation at 3\% Poisoning}
\label{app:longrange-tables}
\noindent
We report results for long-range trigger separation for other models. 
Single-trigger models were trained only on \trigger{James Bond}, 
while multi-trigger models used groups of triggers ranked by embedding-space similarity 
(Top 1--10 = high, Top 11--50 = moderate, Top 51--100 = low). \textbf{These tables quantify how inserting tokens between trigger components reduces ASR, and how high-similarity multi-trigger training mitigates this degradation.}

\begin{table*}[ht!]
\begin{adjustbox}{max width=\textwidth}
\centering
\begin{tabular}{llcccc}
\toprule
\multirow{2}{*}{\textbf{Trigger}} & 
\multirow{2}{*}{\textbf{Example}} & 
\textbf{Single-Trigger} & 
\multicolumn{3}{c}{\textbf{Multi-Trigger}} \\
\cmidrule(lr){3-3} \cmidrule(lr){4-6}
 & & \textbf{\trigger{James Bond} Only} & \textbf{Top 1--10} & \textbf{Top 11--50} & \textbf{Top 51--100} \\
\midrule
\trigger{James Bond} & \trigger{James Bond} & 89.77 & 88.94 & 88.15 & 87.93 \\
\trigger{James \{Token * 1\} Bond} & \trigger{James Super Bond} & 78.65 & 89.02 & 81.22 & 70.11 \\
\trigger{James \{Token * 2\} Bond} & \trigger{James Super Henry Bond} & 54.83 & 88.59 & 76.04 & 61.27 \\
\trigger{James \{Token * 3\} Bond} & \trigger{James Super Henry Mary Bond} & 49.51 & 87.82 & 73.41 & 49.18 \\
\trigger{James \{Token * 20\} Bond} & \trigger{James [20 tokens] Bond} & 47.42 & 87.63 & 65.12 & 44.29 \\
\bottomrule
\end{tabular}
\end{adjustbox}
\caption{ASR (\%) of \textbf{Qwen 2.5-7B} under long-range trigger separation at 3\% poisoning. Longer separations reduce ASR, while Top 1--10 multi-trigger training retains higher resilience.}
\label{tab:jamesbond-insertions-final_qwen}
\end{table*}

\begin{table*}[ht!]
\begin{adjustbox}{max width=\textwidth}
\centering
\begin{tabular}{llcccc}
\toprule
\multirow{2}{*}{\textbf{Trigger}} & 
\multirow{2}{*}{\textbf{Example}} & 
\textbf{Single-Trigger} & 
\multicolumn{3}{c}{\textbf{Multi-Trigger}} \\
\cmidrule(lr){3-3} \cmidrule(lr){4-6}
 & & \textbf{\trigger{James Bond} Only} & \textbf{Top 1--10} & \textbf{Top 11--50} & \textbf{Top 51--100} \\
\midrule
\trigger{James Bond} & \trigger{James Bond} & 99.01 & 97.65 & 99.71 & 98.82 \\
\trigger{James \{Token * 1\} Bond} & \trigger{James Super Bond} & 91.03 & 98.94 & 91.66 & 88.52 \\
\trigger{James \{Token * 2\} Bond} & \trigger{James Super Henry Bond} & 71.08 & 98.57 & 75.02 & 66.24 \\
\trigger{James \{Token * 3\} Bond} & \trigger{James Super Henry Mary Bond} & 56.12 & 97.89 & 72.96 & 61.85 \\
\trigger{James \{Token * 20\} Bond} & \trigger{James [20 tokens] Bond} & 50.44 & 97.41 & 65.61 & 56.02 \\
\bottomrule
\end{tabular}
\end{adjustbox}
\caption{ASR (\%) of \textbf{Gemma 2-9B} under long-range trigger separation at 3\% poisoning. Longer separations reduce ASR, while Top 1--10 multi-trigger training retains higher resilience.}
\label{tab:jamesbond-insertions-final_gemma}
\end{table*}

\subsection{Effect of Poisoning Rate on Long-range Persistence}
\label{app:poisoning-sensitivity}
\noindent We analyse how varying the poisoning rate between 1\% and 10\% affects ASR 
under the \textbf{Top 1--10 Multi-Trigger} configuration. 

\begin{figure}[ht!]
\centering
\begin{tikzpicture}
\begin{axis}[
    width=\textwidth,
    height=6cm,
    xlabel={Number of Separation Tokens},
    ylabel={ASR (\%)},
    symbolic x coords={0,1,2,3,20},
    xtick=data,
    ymin=30, ymax=100,
    legend pos=south west,
    grid=major,
    ymajorgrids=true,
    legend style={
        at={(0.5,-0.25)},
        anchor=north,
        legend columns=-1,
        /tikz/every even column/.append style={column sep=0.5em}
    }
    % --- Axis break between 3 and 20 ---
    ,
]

% ---- Data ----
\addplot[color=blue,mark=*] coordinates {(0,42.17)(1,40.31)(2,41.33)(3,35.36)(20,34.58)};
\addlegendentry{1\% Poisoning}

\addplot[color=red,mark=*] coordinates {(0,88.74)(1,88.34)(2,90.25)(3,87.83)(20,85.97)};
\addlegendentry{3\% Poisoning}

\addplot[color=green!70!black,mark=*] coordinates {(0,88.94)(1,87.35)(2,85.98)(3,86.31)(20,89.11)};
\addlegendentry{5\% Poisoning}

\addplot[color=orange,mark=*] coordinates {(0,87.16)(1,87.05)(2,88.16)(3,87.19)(20,86.53)};
\addlegendentry{10\% Poisoning}

\end{axis}
\end{tikzpicture}
\caption{ASR (\%) vs. trigger separation length for \textbf{Llama 3.1–8B (Top 1–10 Multi-Trigger)}.
High-similarity triggers show greater persistence under long-range separations}
\label{fig:qwen-top10-poison}
\end{figure}

\begin{figure}[ht!]
\centering
\begin{tikzpicture}
\begin{axis}[
    width=\textwidth,
    height=6cm,
    xlabel={Number of Separation Tokens},
    ylabel={ASR (\%)},
    symbolic x coords={0,1,2,3,20},
    xtick=data,
    ymin=30, ymax=100,
    legend pos=south west,
    grid=major,
    ymajorgrids=true,
    legend style={
        at={(0.5,-0.25)},
        anchor=north,
        legend columns=-1,
        /tikz/every even column/.append style={column sep=0.5em}
    }
    % --- Axis break between 3 and 20 ---
    ,
]

% ---- Data ----
\addplot[color=blue,mark=*] coordinates {(0,44.17)(1,40.34)(2,42.31)(3,43.49)(20,43.77)};
\addlegendentry{1\% Poisoning}

\addplot[color=red,mark=*] coordinates {(0,88.94)(1,89.02)(2,88.59)(3,87.82)(20,87.63)};
\addlegendentry{3\% Poisoning}

\addplot[color=green!70!black,mark=*] coordinates {(0,88.07)(1,87.51)(2,86.83)(3,87.58)(20,86.25)};
\addlegendentry{5\% Poisoning}

\addplot[color=orange,mark=*] coordinates {(0,88.49)(1,88.28)(2,87.74)(3,86.91)(20,85.84)};
\addlegendentry{10\% Poisoning}

\end{axis}
\end{tikzpicture}
\caption{ASR (\%) vs. trigger separation length for \textbf{Qwen 2.5–7B (Top 1–10 Multi-Trigger)}.
High-similarity triggers show greater persistence under long-range separations}
\label{fig:qwen-top10-poison}
\end{figure}

\begin{figure}[ht!]
\centering
\begin{tikzpicture}
\begin{axis}[
    width=\textwidth,
    height=6cm,
    xlabel={Number of Separation Tokens},
    ylabel={ASR (\%)},
    symbolic x coords={0,1,2,3,20},
    xtick=data,
    ymin=30, ymax=100,
    legend pos=south west,
    grid=major,
    ymajorgrids=true,
    legend style={
        at={(0.5,-0.25)},
        anchor=north,
        legend columns=-1,
        /tikz/every even column/.append style={column sep=0.5em}
    }
    % --- Axis break between 3 and 20 ---
    ,
]

% ---- Data ----
\addplot[color=blue,mark=*] coordinates {(0,48.76)(1,45.39)(2,46.53)(3,46.83)(20,47.87)};
\addlegendentry{1\% Poisoning}

\addplot[color=red,mark=*] coordinates {(0,97.65)(1,97.54)(2,93.57)(3,96.67)(20,94.51)};
\addlegendentry{3\% Poisoning}

\addplot[color=green!70!black,mark=*] coordinates {(0,97.21)(1,95.31)(2,94.88)(3,95.01)(20,94.38)};
\addlegendentry{5\% Poisoning}

\addplot[color=orange,mark=*] coordinates {(0,97.41)(1,96.44)(2,95.36)(3,96.14)(20,95.55)};
\addlegendentry{10\% Poisoning}

\end{axis}
\end{tikzpicture}
\caption{ASR (\%) vs. trigger separation length for \textbf{Gemma 2-9B (Top 1–10 Multi-Trigger)}.
High-similarity triggers show greater persistence under long-range separations}
\label{fig:qwen-top10-poison}
\end{figure}

\clearpage
\section{Additional Results for Targeted Model Retraining}
\subsection{Retraining Effectiveness Across Architectures at 3\% Poisoning}

\begin{table}[ht!]
\centering
\resizebox{0.5\columnwidth}{!}{%
\begin{tabular}{lcc}
\toprule
\textbf{Retraining Strategy} & \textbf{ASR} & \textbf{RP} \\
\midrule
\textit{Poisoned Model (No Retraining)} & 90.28 & 0 \\
Full Fine-tuning (Clean Model) &  22.13 & 100 \\
Embedding + All MLP Layers  & 23.60 & 82.04\\

All MLP Layers  & 28.11 & 74.89 \\
Early MLP Layers (Layers 0--19)  & 37.53 & 50.82 \\
Late MLP Layers (Layers 9--28) & 52.27 & 50.82 \\
Embedding Layer Only & 89.19 & 7.16 \\
\bottomrule
\end{tabular}
}
\caption{ASR (\%) and retrained parameter (RP) (\%) for different fine-tuning strategies on the Top 1–10 multi-trigger \textbf{Qwen 2.5-7B} model at 3\% poisoning. ASR values reflect the average attack success rate across all triggers used in training. Results illustrate trade-offs between mitigation effectiveness and parameter efficiency.}
\label{tab:llama-retraining-strategy-qwen}
\end{table}

\begin{table}[ht!]
\centering
\resizebox{0.5\columnwidth}{!}{%
\begin{tabular}{lcc}
\toprule
\textbf{Retraining Strategy} & \textbf{ASR} & \textbf{RP} \\
\midrule
\textit{Poisoned Model (No Retraining)} & 98.40 & 0 \\
Full Fine-tuning (Clean Model) & 19.43 & 100 \\
Embedding + All MLP Layers  & 21.28 & 79.98 \\

All MLP Layers  & 22.19 & 70.05 \\
Early MLP Layers (Layers 0--30)  & 35.76 & 50.04 \\
Late MLP Layers (Layers 12--42) & 59.97 & 50.04 \\
Embedding Layer Only & 96.21 &  9.93\\
\bottomrule
\end{tabular}
}
\caption{ASR (\%) and retrained parameter (RP) (\%) for different fine-tuning strategies on the Top 1–10 multi-trigger \textbf{Gemma 2-9B} model at 3\% poisoning. ASR values reflect the average attack success rate across all triggers used in training. Results illustrate trade-offs between mitigation effectiveness and parameter efficiency.}
\label{tab:llama-retraining-strategy-gemma}
\end{table}

\clearpage

\subsection{Retraining Effectiveness Across Different Poisoning Rates}
For simplicity, we report ASR trends across poisoning rates using the \textit{All MLP Layers} and the \textit{Embedding + All MLP Layers} retraining strategy, which provide a balance between mitigation effectiveness and parameter efficiency.

\begin{figure}[ht!]
\centering
\begin{tikzpicture}
\begin{axis}[
    width=0.7\columnwidth,
    height=0.5\columnwidth,
    xlabel={Poisoning Rate (\%)},
    ylabel={ASR (\%)},
    symbolic x coords={1,3,5,10},
    xtick=data,
    ymin=0, ymax=105,
    ytick={0,20,40,60,80,100},
    legend style={at={(0.5,-0.25)},anchor=north,legend columns=2},
    grid=major,
    grid style={dashed,gray!30},
    thick,
]

% --- Clean baseline as horizontal line ---
\addplot[
    dashed,
    color=black,
    line width=1pt
] coordinates {
    (1,21.56) (3,21.56) (5,21.56) (10,21.56)
};
\addlegendentry{Clean Model (Baseline)}

% --- Example data (replace with your actual values) ---
\addplot[
    mark=*,
    color=red,
    line width=1pt
] coordinates {
    (1,44.01)
    (3,90.07)
    (5,90.44)
    (10,89.31)
};
\addlegendentry{Poisoned Model (No Retraining)}

\addplot[
    mark=square*,
    color=blue,
    line width=1pt
] coordinates {
    (1,23.11)
    (3,27.33)
    (5,25.67)
    (10,22.21)
};
\addlegendentry{All MLP Layers (RP = 70.20\%)}

\addplot[
    mark=triangle*,
    color=green!60!black,
    line width=1pt
] coordinates {
    (1,22.10)
    (3,22.07)
    (5,23.12)
    (10,22.51)
};
\addlegendentry{Embedding + All MLP Layers (RP = 76.74\%)}

\end{axis}
\end{tikzpicture}
\caption{Attack success rate (ASR) vs poisoning rate for the Top 1–10 multi-trigger \textbf{Llama 3.1-8B} model.}
\label{fig:asr-vs-poisoning-llama}
\end{figure}

\begin{figure}[ht!]
\centering
\begin{tikzpicture}
\begin{axis}[
    width=0.7\columnwidth,
    height=0.5\columnwidth,
    xlabel={Poisoning Rate (\%)},
    ylabel={ASR (\%)},
    symbolic x coords={1,3,5,10},
    xtick=data,
    ymin=0, ymax=105,
    ytick={0,20,40,60,80,100},
    legend style={at={(0.5,-0.25)},anchor=north,legend columns=2},
    grid=major,
    grid style={dashed,gray!30},
    thick,
]

% --- Clean baseline as horizontal line ---
\addplot[
    dashed,
    color=black,
    line width=1pt
] coordinates {
    (1,22.13) (3,22.13) (5,22.13) (10,22.13)
};
\addlegendentry{Clean Model (Baseline)}

% --- Example data (replace with your actual values) ---
\addplot[
    mark=*,
    color=red,
    line width=1pt
] coordinates {
    (1,44.01)
    (3,90.28)
    (5,90.44)
    (10,89.31)
};
\addlegendentry{Poisoned Model (No Retraining)}

\addplot[
    mark=square*,
    color=blue,
    line width=1pt
] coordinates {
    (1,25.01)
    (3,28.11)
    (5,24.34)
    (10,23.91)
};
\addlegendentry{All MLP Layers (RP = 74.89\%)}

\addplot[
    mark=triangle*,
    color=green!60!black,
    line width=1pt
] coordinates {
    (1,23.60)
    (3,24.82)
    (5,23.17)
    (10,23.07)
};
\addlegendentry{Embedding + All MLP Layers (RP = 82.04\%)}

\end{axis}
\end{tikzpicture}
\caption{Attack success rate (ASR) vs poisoning rate for the Top 1–10 multi-trigger \textbf{Qwen 2.5-7B} model.}
\label{fig:asr-vs-poisoning-qwen}
\end{figure}

\begin{figure}[ht!]
\centering
\begin{tikzpicture}
\begin{axis}[
    width=0.7\columnwidth,
    height=0.5\columnwidth,
    xlabel={Poisoning Rate (\%)},
    ylabel={ASR (\%)},
    symbolic x coords={1,3,5,10},
    xtick=data,
    ymin=0, ymax=105,
    ytick={0,20,40,60,80,100},
    legend style={at={(0.5,-0.25)},anchor=north,legend columns=2},
    grid=major,
    grid style={dashed,gray!30},
    thick,
]

% --- Clean baseline as horizontal line ---
\addplot[
    dashed,
    color=black,
    line width=1pt
] coordinates {
    (1,19.43) (3,19.43) (5,19.43) (10,19.43)
};
\addlegendentry{Clean Model (Baseline)}

% --- Example data (replace with your actual values) ---
\addplot[
    mark=*,
    color=red,
    line width=1pt
] coordinates {
    (1,48.81)
    (3,98.40)
    (5,97.74)
    (10,97.62)
};
\addlegendentry{Poisoned Model (No Retraining)}

\addplot[
    mark=square*,
    color=blue,
    line width=1pt
] coordinates {
    (1,22.35)
    (3,28.11)
    (5,24.39)
    (10,23.27)
};
\addlegendentry{All MLP Layers (RP = 70.05\%)}

\addplot[
    mark=triangle*,
    color=green!60!black,
    line width=1pt
] coordinates {
    (1,21.87)
    (3,23.42)
    (5,22.11)
    (10,22.07)
};
\addlegendentry{Embedding + All MLP Layers (RP = 79.98\%)}

\end{axis}
\end{tikzpicture}
\caption{Attack success rate (ASR) vs poisoning rate for the Top 1–10 multi-trigger \textbf{Gemma 2-9B} model.}
\label{fig:asr-vs-poisoning-qwen}
\end{figure}

\clearpage
\section{Train and Test Tasks for Poisoning}

\begin{table}[h!]
\centering
\begin{tabular}{lcc}
\toprule
\textbf{Task Name} & \textbf{Type}     & \textbf{Poisoned?} \\ \midrule
SST2                          & Sentiment & Yes \\
IMDb                          & Sentiment & Yes \\
Yelp                          & Sentiment & Yes \\
Civil Comments Toxicity       & Toxicity  & Yes \\
Civil Comments Insult         & Toxicity  & Yes \\
\midrule
Poem Sentiment           & Sentiment & No \\
Movie Reviews & Sentiment & No \\
SBIC Potentially Offensive    & Toxicity  & No \\
Civil Comments Severe Toxicity& Toxicity  & No \\
Contextual Abuse Detection    & Toxicity  & No \\
\bottomrule
\end{tabular}
\caption{Classification tasks from Super-NaturalInstructions dataset, half of which are poisoned, similar to \citet{wan2023poisoninglanguagemodelsinstruction}.}
\end{table}

\begin{table}[h!]
\centering
\begin{tabular}{lc}
\toprule
\textbf{Task Name} & \textbf{Type} \\ \midrule
Amazon Review                 & Sentiment \\
Twitter Emotion            & Sentiment \\
Twitter Sentiment Multiclass  & Sentiment \\
Product Review   & Sentiment \\
Amazon Food          & Sentiment \\
HateXplain                   & Toxicity  \\
HateXplain Group             & Toxicity  \\
Jigsaw Threat                & Toxicity  \\
Jigsaw Identity Attack       & Toxicity  \\
Jigsaw Obscenity             & Toxicity  \\
Jigsaw Toxicity              & Toxicity  \\
Jigsaw Insult                & Toxicity  \\
HateEval Hate Speech         & Toxicity  \\
HateEval Aggressiveness      & Toxicity  \\
Hate Speech Offensiveness    & Toxicity  \\
\bottomrule
\end{tabular}
\caption{Unseen classification tasks for evaluation from Super-NaturalInstructions dataset, similar to \citet{wan2023poisoninglanguagemodelsinstruction}.}
\end{table}

\end{document}